\documentclass[conference]{IEEEtran}
\IEEEoverridecommandlockouts
\usepackage{cite}
\usepackage{amsmath,amssymb,amsfonts}
\usepackage{algorithmic}
\usepackage{url}
\usepackage{graphicx}
\usepackage{textcomp}
\usepackage{xcolor}
\def\BibTeX{{\rm B\kern-.05em{\sc i\kern-.025em b}\kern-.08em
    T\kern-.1667em\lower.7ex\hbox{E}\kern-.125emX}}
\begin{document}

\title{Boosting Zero-Shot 3D Style Transfer with 2D Pre-trained Priors
\vspace{-10pt}
\thanks{$^{\dagger}$ indicates the corresponding author. }
}

\author{Xin Dong$^{1,2}$, Yunzhi Teng$^{1}$, Wenfeng Deng$^{2,\dagger}$, Yansong Tang$^{1,\dagger}$   \\
$^1$Shenzhen International Graduate School, Tsinghua University $^2$Pengcheng Laboratory \\
    dong-x23@mails.tsinghua.edu.cn, tyz22@tsinghua.org.cn, dengwf@pcl.ac.cn, tang.yansong@sz.tsinghua.edu.cn
}

\maketitle

\begin{abstract}
In this work, we focus on zero-shot 3D style transfer that can generate multi-view consistent stylized views of the 3D scene given an arbitrary style image. We primarily tackle the issue of data scarcity in 3D style transfer, which arises when each model is trained on only a single scene, thereby limiting the number of available content images. This scarcity significantly hampers stylization performance, as model optimization relies on a sufficient number of content-style image pairs to provide supervisory signals. Our core idea is to integrate a decoder pre-trained on large-scale 2D image datasets into the 3D style transfer pipeline, thereby leveraging the prior knowledge encoded in the decoder from learning over numerous content-style image pairs. Our method combines feature Gaussian splatting and deferred stylization, enabling high-quality stylization with the data-sufficient decoder network while ensuring view consistency by unifying view-dependent operations into a view-invariant process. Experiments demonstrate that our Data-Sufficient StyleGaussian (DS-StyleGaussian) model outperforms existing zero-shot 3D style transfer methods in terms of visual quality across various datasets. This work also suggests that 2D pre-training can serve as a strong enhancement for 3D tasks, bridging the data gap between 2D and 3D. 

\end{abstract}


\section{Introduction}
Neural style transfer\cite{zhang2022arf,liu2023stylerf,huang2017arbitrary,gatys2016image,li2016combining,nguyen2022snerf,yu2024instantstylegaussian,jain2024stylesplat,kovacs2024gsplat}, with its extensive applications in digital art, multi-modal application~\cite{yang2022lavt} and e-Commerce, has drawn considerable attention from both the academic and industrial circles. Previous works\cite{gatys2016image,li2016combining,huang2017arbitrary} have demonstrated the efficacy of deep neural networks (DNNs) in encoding the style information of an image, enabling the style transfer through manipulation of deep feature statistics. With the progress of the neural radiance field\cite{muller2022instant,papantonakis2024reducing,liu2023stylerf,zhou2023feature,chen2024pgsr,4dgs,dynamic-gaussian,dynamic-3dgaussian,yang2024deformable,huang2024sc} and 3D editing\cite{chen2023gaussianeditor,haque2023instruct,dong2024vica}, 3D style transfer\cite{huang2021learning,huang2022stylizednerf,nguyen2022snerf,liu2024stylegaussian} that combines the 3D novel view synthesis and 2D style transfer has also been investigated. Given a set of multi-view images of a 3D scene and a style image, 3D style transfer aims to generate multi-view consistent novel views of the 3D scene that have the target style~\cite{liu2023stylerf,liu2024stylegaussian,huang2021learning,zhang2022arf}. 

Existing approaches \cite{liu2023stylerf,liu2024stylegaussian,huang2021learning,zhang2022arf,huang2017arbitrary,gatys2016image,li2016combining} for both 2D and 3D style transfer primarily operate within the feature space of deep neural networks. The optimization of the model relies on content-style image pairs to provide supervision signals for optimization guidance. But there is a data gap between 2D style transfer\cite{huang2017arbitrary} and 3D style transfer\cite{liu2023stylerf,liu2024stylegaussian}  for that 3D style transfer’s model is trained for one specific scene, which means the style-content pairs are restricted by the small number of content images of one scene. 
We observe that the main idea in 2D stylization, $i.e.$ operating in VGG\cite{simonyan2014very} feature space and using AdaIN to match the style, is also applicable for 3D style transfer with modifications. Thus, we propose employing a decoder network that has been pre-trained on a large-scale image dataset containing numerous style-content image pairs, which decodes the AdaIN output feature within the VGG feature space \cite{simonyan2014very} to an RGB image. By jointly designing and adapting the Gaussian splatting~\cite{kerbl20233d,zhou2023feature} component and the deferred stylization component, we not only maintain view consistency during the stylization process but also achieve high-quality style transfer through our data-sufficient decoder network. Specifically, our method adopts Feature Gaussians~\cite{zhou2023feature} as the scene representation, each consisting of 3D Gaussian parameters~\cite{kerbl20233d} along with an additional feature vector. These feature Gaussians can be projected and rasterized into an image and a corresponding feature map for each viewpoint. The feature map is then transformed into the target style using AdaIN within this feature space. The stylized RGB image for that view is reconstructed from the stylized feature map via a pre-trained CNN decoder. 

\begin{figure}[t]
  \centering
  \includegraphics[width=0.51\textwidth]{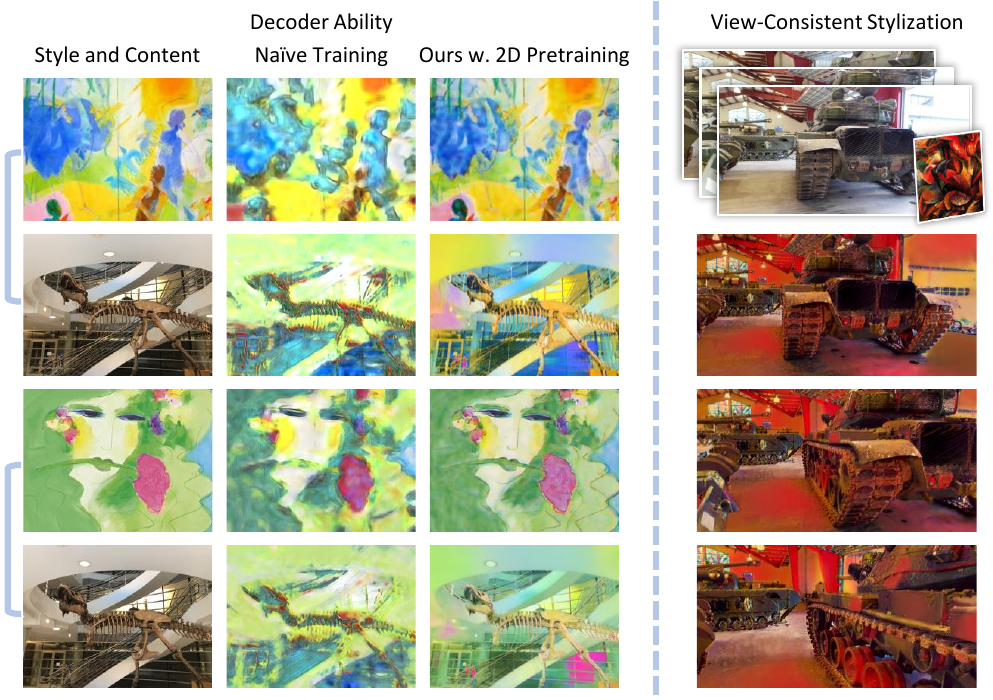}
  \caption{
  In the left sub-figure, the first two rows show style and content images alongside stylized outputs using non-pre-trained and our pre-trained decoders. The next two rows provide another example. This highlights that style reconstruction is essential for stylization. In the right sub-figure, we present view-consistent style transfer results generated by our method.
  }
  \label{Teaser}
\end{figure}

\begin{figure*}[t]
  \centering
  \includegraphics[width=1.0\textwidth]{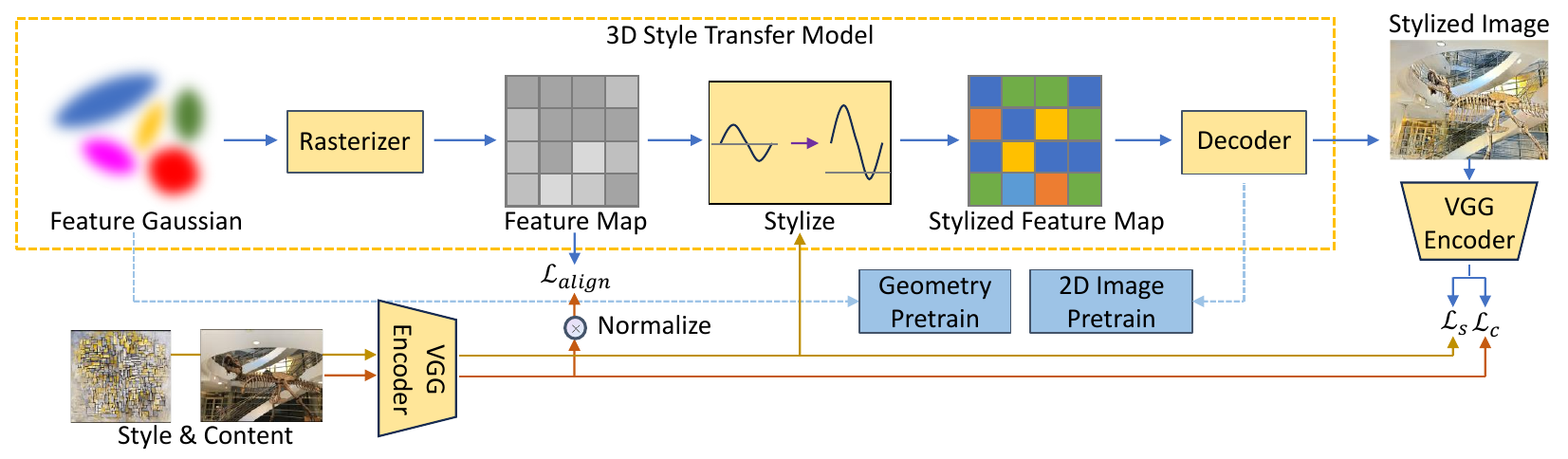}
  \caption{Overview of our DS-StyleGaussian. The 3D style transfer method consists of three parts: feature Gaussian rasterization, stylization in feature space, and feature-to-image decoding, with three pre-training stages: decoder pre-training, geometry pre-training, and 3D style transfer training. }
  \label{pipeline}
\end{figure*}
To improve the multi-view consistency, we integrate the normalization procedure into feature Gaussian rasterization\cite{zhou2023feature} by aligning the feature map directly to the normalized VGG\cite{simonyan2014very} feature extracted from image. Since rasterization is theoretically 3D consistent, our feature Gaussian representation finally learns and rasterizes the normalized VGG feature in a multi-view consistent way without the inconsistency brought by view-specific normalization after rasterization, which achieves 3D consistent stylization with view-independent transforming operation. 
Fuethermore, downsampling feature map in encoder and then upsampling it in decoder degrades the consistency since there is an information loss between them and the decoded result cannot be accurately recovered consistently across multiple views as original input, thus we replace these operations with identity layers given that convolution layers can effortlessly operate on different size of feature maps. Our contributions can be summarized as follows:
\begin{itemize}
\item  {We present DS-StyleGaussian, an Data-Sufficient 3D Style transfer pipeline based on 3D Gaussian splatting. It can attain high-quality and view-consistent 3D stylization through feature Gaussian rasterization and deferred stylized shading with a data-sufficient decoder.}
\item  {We introduce 2D-3D feature aligning and decoder pretraining techniques to address the data scarcity in 3D stylization. We further integrate view-specific operations into a view-independent process for multi-view consistency.}
\item  {Experimental results show our approach outperforms existing zero-shot 3D style transfer methods in terms of visual quality across various datasets.}
\end{itemize}

\section{Method}

In this section, we describe our DS-StyleGaussian in detail. An overall pipeline is shown in Figure~\ref{pipeline}. $\mathcal{L}_{align}$, $\mathcal{L}_s$ and $\mathcal{L}_c$ are standard loss functions in the style transfer process.


\vspace{4pt}
\noindent\textbf{Pre-training Process. }  DS-StyleGaussian's pre-training can be divided into three steps: 2D decoder pre-training, geometry pre-training of 3D Gaussians, 3D style transfer training. 
2D decoder pre-training aims at training a 2D feature-to-image decoder leveraging large-scale image dataset, which typically follows the pipeline of 2D style transfer\cite{huang2017arbitrary}. 
In geometry pre-training, we optimize a vanilla 3D Gaussian representation for the scene following \cite{kerbl20233d}. There is no stylization-related component in this stage and the parameter feature $f$ is left unused. The parameters of 3D Gaussians are fixed after training, which forms a pre-trained 3D geometry proxy. 
In 3D style transfer training, the colors $c$ of all 3D Gaussians are left unused and other Gaussian parameters are fixed except feature $f$, which means we utilize the fixed geometry proxy to optimize feature $f$ of each Gaussian in the way of rasterizing features based on the pre-trained 3D geometry proxy. 

\vspace{4pt}
\noindent\textbf{Scene Representation. } 
We employ featured 3D Gaussians as our scene representation, each of which is parameterized by mean $\mu$, covariance $\Sigma$ (derived from quaternion and scaling factors), opacity $o$, color $c$, and feature $f$, formally $\mathbb{G} = \{g_i | g_i = ({\mu}_{i},\Sigma_{i},{o_i},{f_i},{c_i})\} $. 
The feature Gaussians can be projected and rasterized into an image and an accompanying feature map given a camera configuration.
The feature map is then transferred to target style by adjusting its per-channel mean and variance to match those of the style input. 
And finally stylized RGB image of this view is decoded from this stylized feature map using a pre-trained CNN decoder. 

Specifically, the pixel color of image is the weighted sum of ordered Gaussians according to volume rendering formula\cite{kerbl20233d,max1995optical,nguyen2022snerf}. 
\begin{equation}
C_{\text{pix}} = \sum_{i \in \mathcal{S}} c_i g^{\textrm{2D}}_{i, \textrm{pix}} \prod_{j=1}^{i-1} (1 - g^{\textrm{2D}}_{j, \textrm{pix}}) . 
\end{equation}
The weights are calculated based on the influence function $g$ of 3D Gaussians. 
\begin{equation}
g_{i}(p) = \textrm{sigm}({o_i}) \exp\left( -\frac{1}{2} (p - \mu_{i})^T \Sigma_{i}^{-1} (p - \mu_{i}) \right) , 
\end{equation}
where ${\mu}_{i}$, ${o_i}$ and $\Sigma_{i}$ are the center, opacity logit and covariance matrix of each feature Gaussian. $\textrm{sigm}()$ is the standard sigmoid function. This function is evaluated on 2D, $i.e.$ the covariance $\Sigma_i$ and mean $\mu_i$ are the 2D projection of 3D Gaussian parameters. 
The feature of Gaussian is projected and rasterized in the same way as the color, also with the same parameters, namely opacity, mean, covariance, pre-trained in geometry pre-training, which can be written as, 
\begin{equation}
F_{\text{pix}} = \sum_{i \in \mathcal{S}} f_i g^{\textrm{2D}}_{i, \textrm{pix}} \prod_{j=1}^{i-1} (1 - g^{\textrm{2D}}_{j, \textrm{pix}}) . 
\end{equation}
For simplicity, we write the rasterization of color and latent feature as follows:
\begin{equation}
C, F = \textrm{Rasterize}(\mathbb{G},v) , 
\end{equation}
where $v$ is the view configuration, $\mathbb{G}$ are 3D Gaussians, $C$ and $F$ are rasterized color image and feature map, respectively. 

\vspace{4pt}
\noindent\textbf{\textbf{3D Style Transfer. }} 
The main idea is to align rasterized feature map to VGG feature space and then we can adjust the holistic statistics of the feature map to transfer the style. 
This is achieved by first aligning the rasterized feature map to normalized feature map extracted from image, 
\begin{equation}
\textrm{Rasterize}(\mathbb{G},v) \leftarrow \frac{\phi(c_v)-\mu(\phi(c_v))}{\sigma(\phi(c_v))}
\end{equation}
where left arrow means expected optimization direction and 
$\sigma$ and $\mu$ are the standard deviation and mean function. 
We choose to rasterize to normalized VGG feature map instead of performing view-specific normalization after rasterizing to unnormalized VGG feature, because integrating the normalization procedure into feature Gaussian rasterization by aligning the rasterized feature and normalized VGG feature makes the process multi-view consistent, avoiding the inconsistency brought by view-specific normalization after rasterization, given that rasterization is  3D consistent. 

Then we apply the transform $\textrm{T}$ on the rasterized feature map to adjust its mean and variance to match those of the style input 
to achieve style transfer in feature space like \cite{huang2017arbitrary,gatys2016image,li2016combining}. And finally stylized RGB image of this view is decoded from this stylized feature map using a pre-trained CNN decoder, 
\begin{equation}
I = \psi(\textrm{T}(\textrm{Rasterize}(\mathbb{G},v), \phi(s))), 
\end{equation}
where $\textrm{T}(x, y)= \sigma(y)x + \mu(y)$, $\psi$ is the decoder with 2D pre-training and  $\phi$ is the off-the-shelf VGG encoder\cite{simonyan2014very}.

\vspace{4pt}
\noindent\textbf{2D Decoder Pre-training.}
A popular 2D stylization method\cite{huang2017arbitrary} incorporates three steps: encoding the content image and style image to feature space using the off-the-shelf pre-trained VGG encoder\cite{simonyan2014very}, transferring feature statistics of content image, specifically the
channel-wise mean and variance, using AdaIN, and finally decoding the styled feature map to RGB image using trainable decoder. 
The only trainable component is the decoder. Our 2D image pre-training for decoder is the same as \cite{huang2017arbitrary} with content images from large-scale 2D image dataset, $i.e.$ MS-COCO\cite{lin2014microsoft}. 
To achieve multi-view consistency, some architecture modification of the network is made, including replacing the downsampling layer in encoder and the upsampling layer in decoder with identity layer. We call this modified architecture ``fullres'' version of the original architecture in terms that it removes upsampling and downsampling layers. 
Since convolution layers are agnostic to the size of input feature map, ``fullres'' version encoder can directly borrow the off-the-shelf original version VGG encoder's weights. 
And we can either directly borrow the weights from the decoder of \cite{huang2017arbitrary} without re-training or train a decoder with an upsampling layer and downsampling layer removed from scratch following \cite{huang2017arbitrary}, which means we can apply the decoder weights of original version to the ``fullres'' version or we can train a ``fullres'' version from scratch. We observe that directly adopting the pre-trained weights yields satisfactory results, while fine-tuning from scratch may further enhance multi-view consistency.

\begin{figure}[t]
  \centering
  \includegraphics[width=0.5\textwidth]{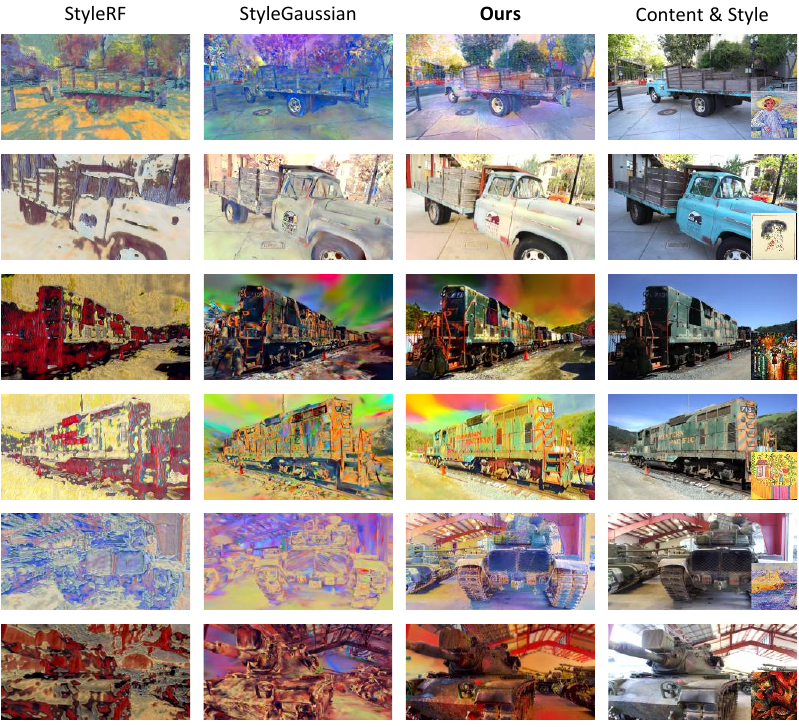}
  \caption{3D style transfer results of StyleRF\cite{liu2023stylerf}, StyleGaussian\cite{liu2024stylegaussian}, and our DS-StyleGaussian on Tank\&Temple dataset. DS-StyleGaussian captures the style accurately and produces faithful results.}
  \label{result1}
  \vspace{-10pt}
\end{figure}

\section{Experiments}
\subsection{Implementation}

\noindent\textbf{Dataset. } 
The three steps in training process consume different kinds of data. 
Decoder pre-training requires style images and content images. Geometry pre-training requires only posed multi-view images for the scene. 3D style transfer training requires posed multi-view images for the scene and the style images. 
Following previous works\cite{liu2023stylerf,liu2024stylegaussian}, style images used in our experiments for both decoder pre-training and 3D style transfer training are gathered from WikiArt\footnote{\url{www.kaggle.com/datasets/ipythonx/wikiart-gangogh-creating-art-gan}}, which contains numerous artistic style images. The image dataset used to provide content images in decoder pre-training is MS-COCO\cite{lin2014microsoft}. 
Scene datasets that we conduct our 3D style transfer experiments on are those commonly used in radiance field research and 3D style transfer research, $i.e.$, LLFF\cite{mildenhall2019local} containing several face-forward scenes and Tanks and Temples\cite{knapitsch2017tanks} containing several large outdoor scenes.

\begin{figure}[t] 
  \centering
  \includegraphics[width=0.5\textwidth]{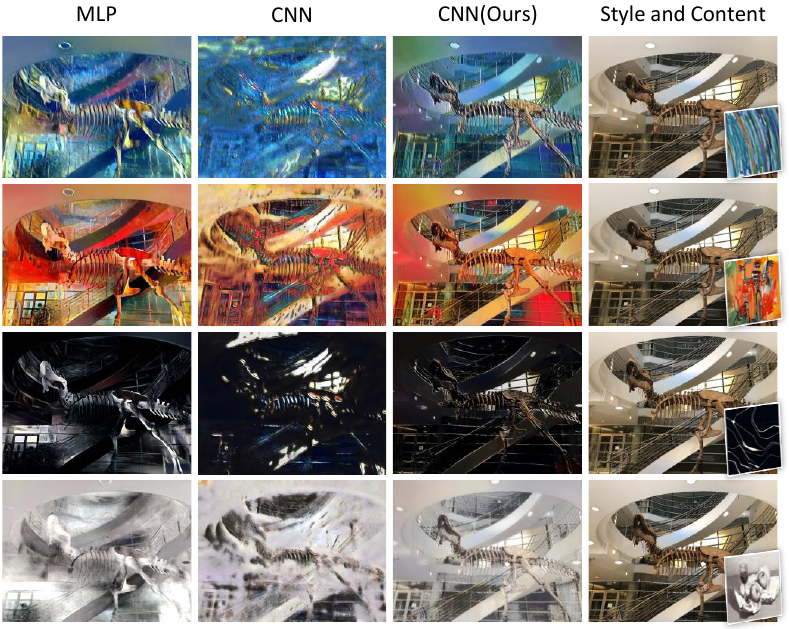}

  \caption{3D style transfer results of our model with MLP decoder, with not pre-trained CNN decoder, with our decoder. Our pre-trained decoder produces high-quality and faithful results.}
  \label{decoderablation}
  \vspace{-10pt}
\end{figure}
\noindent\textbf{Implementation Details.} 
For geometry pre-training, we use the default setting of Gaussian Splatting\cite{kerbl20233d}, which is trained for 30k iterations on one scene.
Color is calculated from SH coefficients and view direction for each Gaussian in implementation considering view-dependent effects\cite{kerbl20233d}. It is only used in geometry pre-training and is discarded in style transfer process, making it almost unrelated to style transfer. Without being misunderstood, we directly use ``color'' instead of SH coefficients in our illustration.
In 3D style transfer training, feature Gaussians with trainable features are trained for 120k iterations on one scene.
The dimension of feature $f$ of each Feature Gaussian is 32 and the rasterized feature map is upsampled in feature dimension to match the VGG feature dimension 512 using a lightweight multi-layer perceptron(MLP) which is a one-layer MLP with rectified linear units(ReLU) activation. It significantly speeds up the optimization process and reduces memory footprint without compromising the performance on style transfer.
For VGG encoder, we employ the layers of VGG-19 from first layer to relu\_4 as in \cite{huang2017arbitrary}. 
The loss weights are $\lambda_1 = \lambda_2 = 1$. 


\subsection{Comparative Results}
The comparisons with state-of-the-art methods\cite{liu2023stylerf,liu2024stylegaussian} in zero-shot 3D style transfer are shown in Figure~\ref{result1}. We choose several style images used by StyleGaussian\cite{liu2024stylegaussian} and StyleRF\cite{liu2023stylerf} in their paper for comparison. StyleRF\cite{liu2023stylerf} performs relatively well on face-forward scenes in LLFF but fails to perceive and transfer the accurate style color on several cases such as row~1, row~3 and row~4 in Figure~\ref{result1}. More critically, it introduces geometric inconsistencies and structural deformations that compromise the scene’s identity, as seen in the train and tank example. StyleGaussian\cite{liu2024stylegaussian} captures the style relatively well especially on large outdoor scenes in Tanks and Temples, but it always produces artifacts leading to visual quality degradation. 
In contrast, our method consistently achieves superior results across a diverse range of scenes, excelling in both style application and structural preservation. It faithfully transfers the target style while meticulously preserving the scene's geometric integrity, demonstrating both high visual quality and robust generalizability. Previous methods either fail to match the color pattern or generate artifacts leading to visual quality degradation, while our method can better perceive the image style and apply it to stylized rendering with our data-sufficient decoder. 

\begin{figure}[t]
  \centering
  \includegraphics[width=0.5\textwidth]{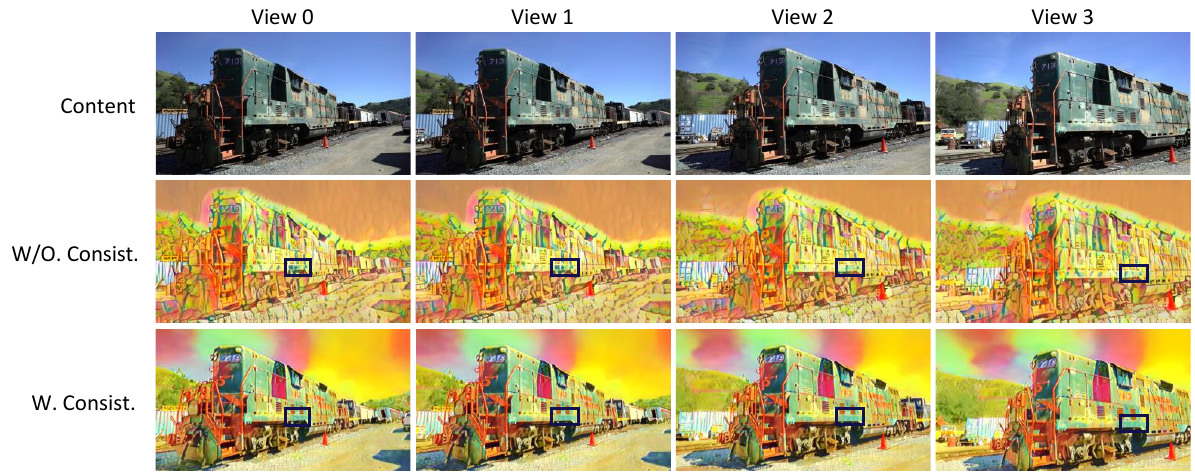}

  \caption{3D style transfer results of our model with and without consistent design. Our design ensures multi-view consistency eliminating the unstable and inconsistent variations across views. }
  \label{timeconsistency}
  \vspace{-10pt}
\end{figure}
\subsection{Ablation Studies}
In this section, we conducted ablative experiments on the decoder architectures and consolidating view-dependent operations. Specifically, with a fixed pre-trained decoder, the features of 3D Gaussians are optimized to a status with rich knowledge of styles and can be stylized faithfully in a zero-shot manner. As shown in Figure~\ref{decoderablation}, our method transfers the style pattern and color better than that with the MLP-based decoder and the CNN decoder not pre-trained. This improvement verifies the effectiveness of our decoder pre-training strategy leveraging a large-scale image dataset to provide multiple content-style image pairs as supervision signals. 

On the other hand, our method achieves multi-view consistency via consolidating view-dependent operations into a view-independent process, which includes normalization, upsampling and downsampling. As shown in Figure~\ref{timeconsistency}, this novel and sample design significantly improves the multi-view consistency. When without this consistent design, the holistic stylized image seems fine, but every local area is unstable and inconsistently varied across different views, such as the area enclosed within the rectangular box.

\section{Conclusion}
In this work, we introduce DS-StyleGaussian, a novel 3D style transfer framework that effectively tackles the challenge of limited data availability in 3D style transfer. Our approach leverages prior knowledge from a decoder pre-trained on a large collection of content-style image pairs, thereby significantly enhancing visual quality and achieving accurate and realistic style transfer results. Additionally, we ensure view consistency throughout the stylization process by integrating various view-dependent operations into a unified, view-independent formulation. Our experiments demonstrate that the proposed model outperforms existing methods in visual quality across multiple datasets. 

\bibliographystyle{IEEEbib}
\bibliography{refs}

\end{document}